\documentclass[pdflatex,sn-mathphys-num]{sn-jnl}% Math and Physical Sciences Numbered 
\usepackage{amssymb}

\usepackage{graphicx}%
\usepackage{multirow}%
\usepackage{amsmath,amssymb,amsfonts}%
\usepackage{amsthm}%
\usepackage{mathrsfs}%
\usepackage[title]{appendix}%
\usepackage[table]{xcolor}%
\usepackage{textcomp}%
\usepackage{manyfoot}%
\usepackage{tcolorbox}
\usepackage{booktabs}%
\usepackage{algorithm}%
\usepackage{algorithmicx}%
\usepackage{algpseudocode}%
\usepackage{listings}%
\usepackage{soul}
\usepackage{enumitem}
\usepackage{hyperref}
\usepackage{comment} 
\usepackage[utf8]{inputenc}
\usepackage{booktabs}      
\usepackage{tabularx}     
\usepackage{float}
\usepackage{subcaption}

\theoremstyle{thmstyleone}%

\theoremstyle{thmstyletwo}%

\theoremstyle{thmstylethree}%

\makeatletter
\def\addressfont{%
  \reset@font
  \fontsize{10bp}{11bp}\selectfont
  \titraggedcenter
}
\makeatother
\begin{document}

\title[]{Responsible Integration of AI in Cancer Genomics: Barriers, Risks, and Pathways to Trustworthy Clinical Translation}

\author*[1]{\fnm{Bahar} \sur{{\.I}lgen}}\email{IlgenB@rki.de}

\author[2]{\fnm{Yiannos} \sur{Tolias}\textsuperscript{\dag}}

\begingroup
\renewcommand\thefootnote{\dag}
\footnotetext{The views expressed are personal, do not necessarily represent the official position of the European Commission.}
\endgroup  

\author[1]{\fnm{Denise} \sur{Kühnert}}

\author[3]{\fnm{Paraskevi} \sur{Papadopoulou}}

\author[4, 5]{\fnm{Magnus} \sur{Westerlund}}

\author[6]{\fnm{Dominik} \sur{Heider}}

\author[1, 7]{\fnm{Katharina} \sur{Ladewig}}

\author[1,8]{\fnm{Georges} \sur{Hattab}}

\affil*[1]{\orgdiv{Centre for Artificial Intelligence in Public Health Research (ZKI-PH)}, \orgname{\\Robert Koch Institute}, \orgaddress{\city{Berlin}, \country{Germany}}\vspace{2pt}}

\affil[2]{\orgname{Directorate-General for Health and Food Safety, }%
\par
\orgname{European Commission},% 
 \orgaddress{\city{ Brussels}, \country{Belgium}}\vspace{2pt}}

\affil[3]{\orgname{Department of Natural Sciences, School of Science and Technology,}%
\par
\orgname{Deree-The American College of Greece},% 
 \orgaddress{\city{ Athens}, \country{Athens}}\vspace{2pt}}

\affil[4]{\orgname{{\AA}bo Akademi University}, \orgaddress{\city{Turku}, \country{Finland}}\vspace{2pt}}

\affil[5]{\orgname{Arcada University of Applied Sciences}, \orgaddress{\city{Helsinki}, \country{Finland}}\vspace{2pt}}

\affil[6]{\orgdiv{Institute of Medical Informatics}, \orgname{University of Münster}, \orgaddress{\city{Münster}, \country{Germany}}\vspace{2pt}}

\affil[7]{\orgdiv{Medical University Lausitz - Carl Thiem (MUL-CT)}, \country{Germany}\vspace{2pt}}

\affil[8]{
\orgname{Department of Mathematics and Computer Science,}%
\par
\orgname{Freie Universität Berlin},% 
 \orgaddress{\city{ Berlin}, \country{Germany}}\vspace{2pt}}

\abstract{
       Artificial intelligence (AI) and natural language processing (NLP) are increasingly used to extract, integrate, and interpret biomedical knowledge relevant to cancer genomics, yet their translation into routine clinical oncology has been comparatively slow. The central challenge is not computational capability alone, but trustworthy integration into clinical workflows. This review examines how NLP and AI support the cancer genomics pipeline, from literature mining and automated variant interpretation to clinical trial matching, knowledge graph construction, and multimodal data integration. We identify four interrelated translational failure domains: evidence inconsistency, explainability and uncertainty, data governance and reproducibility, and interoperability. Rather than considering these challenges in isolation, we take a systems-level view, focusing on their interaction across the translational pathway. We propose a conceptual framework and roadmap for addressing these domains through rigorous validation, uncertainty-aware methods, interoperable infrastructures, regulatory alignment, and human oversight across the AI lifecycle. Progress toward routine clinical use will depend less on further improving model capability than on systematically addressing these interacting failure domains from development through deployment and post-deployment monitoring.
}
 

\maketitle

%\vspace{-12pt}
\section{Introduction}\label{sec1}

  The principal obstacle to clinical artificial intelligence (AI) in cancer genomics is no longer computational capability alone but trustworthy clinical translation. Advances in cancer genomics, next-generation sequencing (NGS), and AI have dramatically expanded the ability to generate and analyse genomic data~\cite{MARDIS2008133, TCGA2013PanCancer, ICGC2010, MacConaill2010, Xu2019}. However, translating these advances into routine clinical oncology continues to present substantial challenges, requiring AI systems that are not only technically robust but also trustworthy, clinically reliable, and suitable for real-world deployment.
  
  This translational challenge is particularly evident in cancer genomics, which has become one of the most data-intensive areas of modern biomedicine. Large-scale sequencing initiatives and routine clinical sequencing generate vast amounts of genomic information~\cite{TCGA2013PanCancer, ICGC2010, Garraway2013Lessons}, while biomedical literature, clinical records, and molecular databases continue to expand rapidly. This creates a fundamental integration challenge: clinically relevant knowledge is distributed across heterogeneous and often unstructured sources that are difficult to interpret, validate, and combine at scale. Conventional approaches increasingly struggle to manage the volume and complexity of modern genomic and biomedical data. Natural language processing (NLP) has therefore become an important component of modern cancer genomics workflows. By extracting and integrating information from biomedical literature, clinical records, genomic databases, and other unstructured sources, these approaches help transform fragmented evidence into clinically relevant knowledge. Applications now span literature mining, automated variant interpretation, clinical decision support, clinical trial matching, knowledge graph construction, biomarker discovery, multimodal data integration, and the extraction of gene–variant–disease relationships~\cite{Xu2019, Lipkova2022MultimodalAI, Waqas2024MultimodalOncologyReview}.
  
  Explainability, rare-variant interpretation, interoperability, data governance, and clinical validation are open problems, and progress on any one does not guarantee progress on the others. The pace of AI development in genomics has outstripped the infrastructure needed to evaluate, communicate, and act on its outputs in real clinical settings. This review builds on the work of the EU Health Policy Platform Thematic Network on "Natural Language Processing for Cancer Genomics", which brought together experts in genomics, oncology, NLP, machine learning, computational linguistics, biomedical ethics, and health policy~\cite{eu2024jointstatement}. The collaboration resulted in a Joint Statement, endorsed by European institutions, outlining scientific, technical, and ethical priorities for the responsible use of AI and NLP in oncology. 
  
\begin{figure*}[t]
\centering
\includegraphics[width=0.92\textwidth]{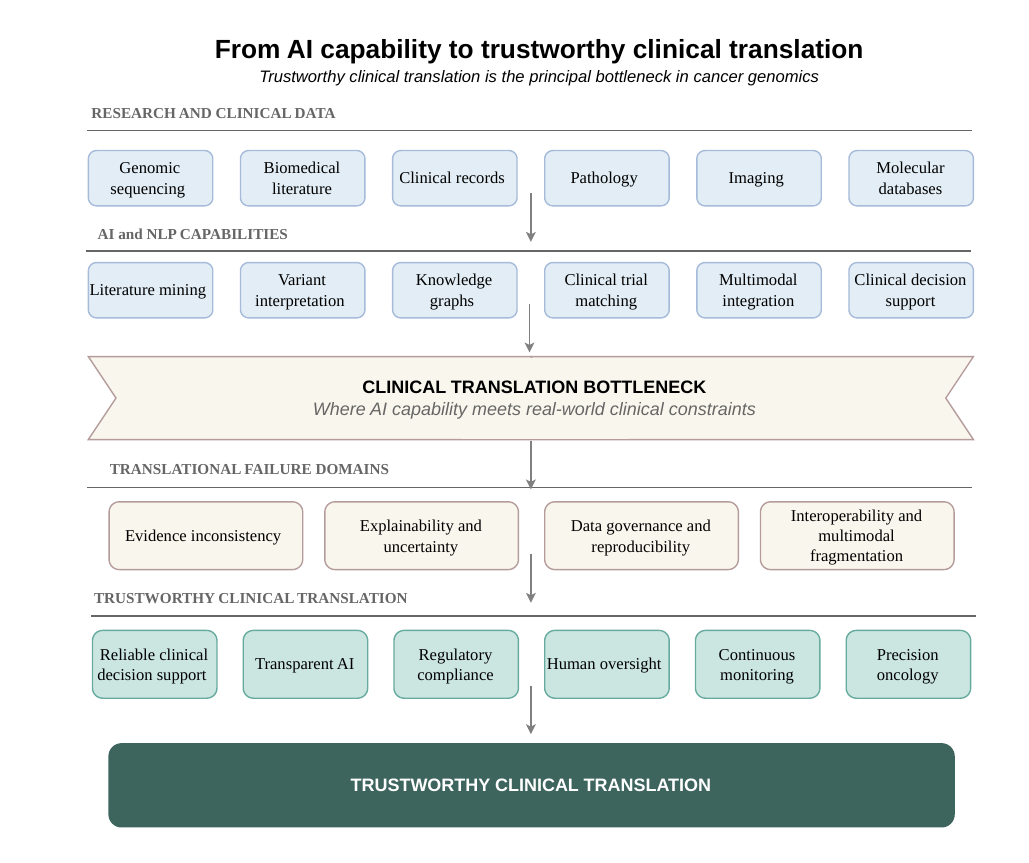}
\caption{
    Conceptual framework illustrating the translational pathway from AI capability to trustworthy clinical translation in cancer genomics. Advances in genomic sequencing, biomedical knowledge resources, and NLP have expanded AI capabilities across evidence synthesis, variant interpretation, knowledge graph construction, multimodal integration, and clinical decision support. Their routine clinical translation, however, is limited by interconnected failure domains, including evidence inconsistency, explainability and uncertainty, data governance, and interoperability. Overcoming these interacting barriers is essential for achieving reliable, transparent, and clinically integrated AI systems in precision oncology.
}
\label{fig:translation_framework}
\end{figure*}

  Building on these observations, this review examines why clinical translation remains challenging despite substantial technological progress. Rather than organizing the field around individual applications or algorithmic advances, we adopt the conceptual framework illustrated in Fig.~\ref{fig:translation_framework} to organize the discussion around the main barriers to clinical integration: evidence inconsistency, explainability and uncertainty, data governance, and interoperability. Accordingly, this review is organized around three questions:

    \begin{itemize}
        \item How is NLP currently used across the cancer genomics pipeline?
        \item Which failure domains limit the trustworthy clinical translation of NLP-enabled AI systems into routine oncology practice?
        \item Which pathways may enable trustworthy and responsible clinical integration?
    \end{itemize}
  
   The following sections address these questions within the proposed translational framework.

  %---------------- FIGURE 1----------------

\section{Failure Points for Trustworthy Clinical Integration}\label{sec2}

\subsection{From Research Utility to Clinical Integration}

     The rapid adoption of AI in cancer genomics has been driven by two parallel developments: the expansion of large-scale genomic sequencing and the growing availability of biomedical knowledge that can be computationally integrated using NLP~\cite{Garraway2013Lessons, TCGA2013PanCancer, ICGC2010}. Together, these developments have increased reliance on NLP to integrate unstructured biomedical knowledge, including scientific literature, clinical narratives, pathology reports, and trial registries, with genomic evidence for clinical interpretation.

    Major international initiatives, including The Cancer Genome Atlas (TCGA), the International Cancer Genome Consortium (ICGC), and the ICGC/TCGA Pan-Cancer Analysis of Whole Genomes (PCAWG) consortium, have generated large-scale multimodal clinicogenomic resources that capture genomic, molecular, pathological, and clinical information across diverse tumour types. These efforts have substantially enabled the development and validation of AI methods for cancer research while simultaneously highlighting the increasing molecular complexity and heterogeneity of cancer, thereby intensifying the need for computational methods capable of integrating genomic findings with biomedical literature, clinical records, and other heterogeneous biomedical data and knowledge sources~\cite{TCGA2013PanCancer, ICGC2010, Gerstung2020Evolution, Garraway2013Lessons, Zitnik2019MLIntegration}.

    As genomic evidence continues to expand across multiple molecular modalities (including somatic mutations, transcriptomics, epigenomics, pathology, and clinical phenotypes) the challenge has progressively shifted from generating genomic data to integrating heterogeneous evidence into clinically actionable knowledge. This transition has positioned NLP and multimodal AI as key components of contemporary clinicogenomic workflows~\cite{Lipkova2022MultimodalAI, Waqas2024MultimodalOncologyReview, Valous2024GraphMLMultiOmics}.

\begin{figure*}[t]
\centering
\includegraphics[width=1\textwidth]{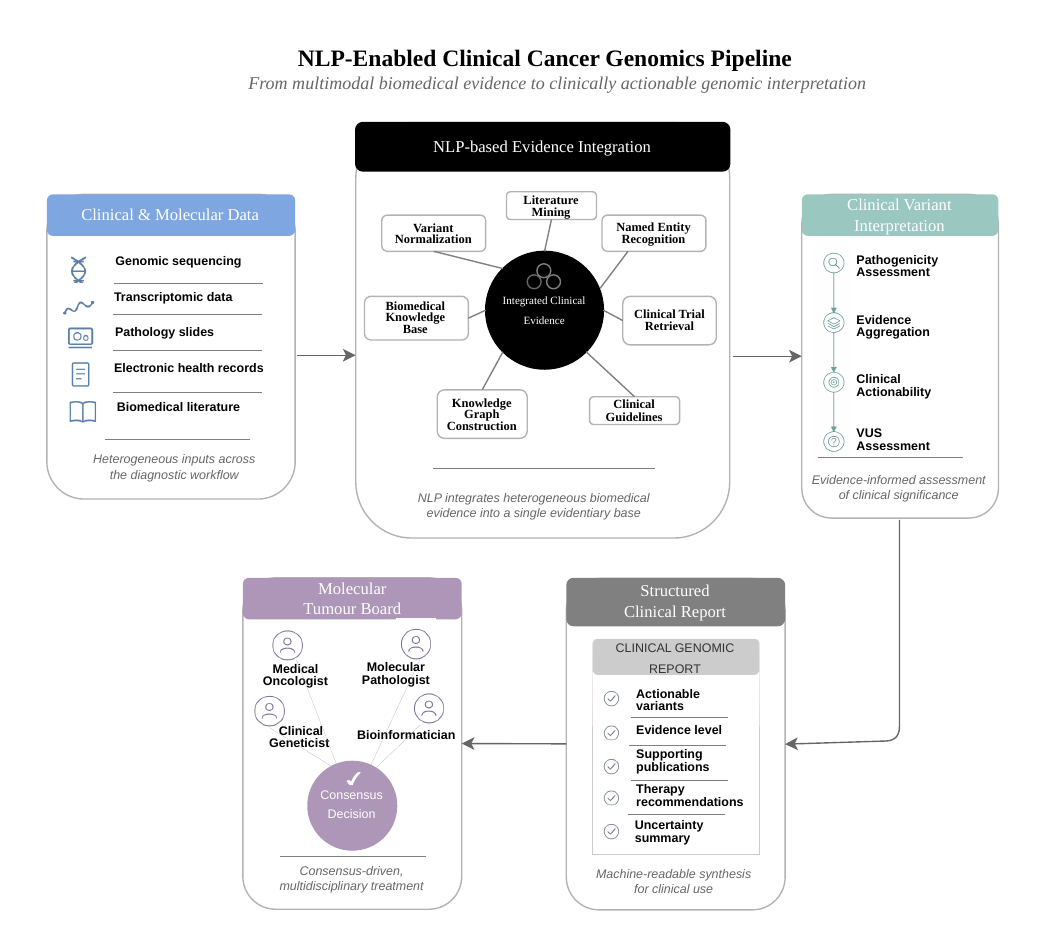}
\caption{
    Overview of the NLP-enabled clinical cancer genomics pipeline. NLP integrates multimodal biomedical evidence (including genomic sequencing, biomedical literature, clinical records, pathology, and molecular databases) to support evidence synthesis, clinical variant interpretation, structured genomic reporting, and multidisciplinary molecular tumour board decision-making. Trustworthy clinical translation requires explainability, human oversight, interoperability, data governance, regulatory compliance, and continuous validation throughout the pipeline.
}
\label{fig:ngs_tumorboard_workflow}
\end{figure*}

    Over the past decade, NLP has demonstrated substantial utility across a wide range of cancer genomics applications. In \emph{literature mining and knowledge extraction}, domain-specific language models such as BioBERT~\cite{lee2020biobert}, SciBERT~\cite{beltagy-etal-2019-scibert}, PubMedBERT~\cite{10.1145/3458754}, and BioGPT~\cite{luo2022biogpt} automate the discovery of variant--disease links and biomarkers from millions of oncology publications, with systems such as LitVar~\cite{10.1093/nar/gky355}, PubTator~\cite{10.1093/nar/gkae235}, and CIViCmine~\cite{Lever2019CIViC} continuously updating curated cancer knowledge bases
    from newly published evidence. In \emph{clinical genomic reporting}, NLP aggregates evidence from biomedical databases, literature, and clinical annotations to support the classification of variants of uncertain significance and to identify actionable mutations for therapeutic decision-making~\cite{Lever2019CIViC,Botsis2022NLPGenomics,lin2019variant2literature}. \emph{Knowledge graph construction} extends this further: NLP-driven entity and relation extraction organises genes, variants, drugs, and phenotypes into structured, queryable networks. Resources such as the LitVar KG, CTKG~\cite{Chen2022ClinicalTrialsKG}, and the Drug--Gene Interaction Database (DGIdb)~\cite{Griffith2013DGIdb} illustrate how such graphs support novel gene--disease discovery, drug--target--biomarker mapping, and explainable clinical decision support~\cite{Zhang2025BiomedicalKG, Cui2025ReviewHKG}. In \emph{cohort selection and trial matching}, NLP extracts eligibility attributes from EHRs, pathology reports, and genomic data to accelerate patient--trial matching and improve representation of rare disease subtypes, integrating live registry feeds from resources such as ClinicalTrials.gov and EUCTR to help maintain up-to-date eligibility assessments~\cite{Klein2022MatchMiner, mazor2025clinical}. Finally, \emph{end-to-end workflows} increasingly combine these components, moving from variant call files through NLP-supported evidence aggregation to structured, human-reviewed reports for molecular tumour boards (Fig.~\ref{fig:ngs_tumorboard_workflow}), while embedding explainability, human-in-the-loop checkpoints, and privacy-preserving deployment throughout the pipeline~\cite{Tiwari2025AIcancer, 10.1007/978-3-031-08999-2_1, Lipkova2022MultimodalAI}. By enabling the extraction and synthesis of evidence from rapidly growing biomedical corpora, NLP has become an important component of precision oncology workflows and a valuable bridge between genomic data and clinical knowledge.

    Despite this broad research utility, many systems demonstrate promising performance in experimental and retrospective settings, yet few achieve sustained clinical deployment. This gap reflects more than limitations in accuracy or efficiency: clinical implementation requires systems to operate reliably within heterogeneous data environments, evolving evidence bases, regulatory constraints, and high-stakes decision-making. Reliability, interpretability, reproducibility, interoperability, governance, and clinician trust are as important as predictive performance, and failures in any of these dimensions may undermine clinical integration.
    
    The key question is thus not only what NLP can do for cancer genomics, but why promising systems often fail to move from research settings into routine care. Addressing this question means moving beyond capability-oriented evaluation to examine the scientific, technical, clinical, and governance conditions required for trustworthy implementation. The following sections outline these translational failure domains and discuss how they can be addressed to support responsible AI integration in precision oncology.

                    %---------------- FIGURE 2----------------

\subsection{Translational Failure Domains}

    In cancer genomics, the trustworthiness of NLP-enabled AI depends on interacting scientific, technical, clinical, and governance-related factors. Accordingly, we organise the remainder of this review around four interrelated translational failure domains:

\begin{itemize}
    \item \textbf{Evidence inconsistency and clinical reliability}: The quality, consistency, and clinical interpretability of the underlying genomic evidence base.
    \item \textbf{Explainability and uncertainty}: Enabling clinicians to understand, verify, and appropriately interpret model outputs.
    \item \textbf{Data governance and reproducibility}: The data standards, provenance, and infrastructure required for reliable and reproducible clinical AI deployment.
    \item \textbf{Interoperability and multimodal fragmentation}: The technical and procedural compatibility required to integrate heterogeneous data across institutions and modalities.
\end{itemize}

    These translational failure domains are closely interconnected rather than independent. Weaknesses in one domain may propagate across subsequent stages of the translational pathway, limiting safe and effective clinical deployment. The following sections examine each domain before outlining practical pathways toward responsible integration.

\section{Evidence Inconsistency and Clinical Reliability}\label{sec3}

\subsection{Variant Interpretation Challenges}

    Accurate variant interpretation is a central challenge in precision oncology. NLP-enabled AI can support the identification of clinically actionable variants by integrating genomic findings with evidence from biomedical literature, clinical records, and curated knowledge bases~\cite{Tiwari2025AIcancer}. However, computational performance alone is insufficient for reliable interpretation.

    NLP systems can also support the assessment and prioritisation of variants of uncertain significance (VUS) by integrating case reports, functional studies, and molecular context. Resources such as CIViC~\cite{Lever2019CIViC}, PubMiner~\cite{Botsis2022NLPGenomics}, and variant2literature~\cite{lin2019variant2literature} illustrate scalable variant--literature integration, while transformer-based models~\cite{lee2020biobert,10.1145/3458754} support context-aware evidence retrieval and biomedical relation extraction. Their reliability, however, is constrained by the underlying evidence: differences in database curation, conflicting findings, incomplete functional validation, and variable evidence quality can produce inconsistent variant classifications across knowledge sources.

    Improved NLP models are unlikely to resolve uncertainty or conflict already present in the biomedical evidence; they may instead retrieve and synthesise it. Trustworthy variant interpretation also requires evidence generation, curation, validation, and maintenance of biomedical knowledge resources.

\subsection{Rare Variants and Evidence Scarcity}

  Rare variants expose an important evidence gap in cancer genomics. Clinical interpretation is often limited less by algorithms than by the lack of biological validation. NGS frequently detects novel or low-frequency variants with incomplete functional evidence, and databases such as ClinVar, gnomAD, and CIViC still leave important gaps, especially for rare variants and underrepresented populations.

  NLP can partially address this challenge by extracting sparse literature evidence, linking related variants, and supporting pathogenicity assessment. Sequence-based approaches can complement this evidence using pretrained protein or DNA foundation models for zero-shot variant-effect prediction~\cite{Alfisi2025DNAFM}. However, model-derived predictions cannot substitute for experimental validation when used for clinical decision-making.

  Orthogonal evidence from proteomics and epigenomics may further support variant assessment, while emerging computational methods, such as deep-set networks for integrating variant annotations~\cite{Clarke2024DeepSet} and knowledge-graph-based approaches for evidence integration and prioritisation~\cite{Ryu2026IDAP}, can improve variant prioritisation and evidence integration. However, these approaches do not eliminate the underlying scarcity of validated biological evidence. Rare variant interpretation thus illustrates a broader limitation of trustworthy clinical AI: computational advances cannot fully compensate for missing biological evidence. Reliable interpretation still requires functional studies, appropriately powered population-scale genomic resources, and systematic clinical validation alongside advances in AI methodology.

\subsection{Functional Validation Gaps}
    Functional validation is a principal bottleneck in trustworthy clinical variant interpretation. Even when NLP-enabled AI systems accurately retrieve and synthesise available evidence, reliable clinical interpretation requires experimental confirmation of biological function, which computational methods alone cannot provide.

    Literature-mining systems such as LitVar~\cite{10.1093/nar/gky355}, PubTator~\cite{10.1093/nar/gkae235}, and CIViCmine~\cite{Lever2019CIViC} retrieve published evidence on variant effects in key cancer genes (e.g., TP53 and BRCA1), while machine-learning models integrate coding and non-coding variants with multi-omics features to predict pathogenicity~\cite{Pepe2025reannotation}. Neither retrieval nor prediction, by itself, establishes functional impact. This distinction is codified in clinical practice: under the ACMG/AMP framework, computational predictions receive supporting-level evidence codes (PP3/BP4), while strong functional evidence codes (PS3/BS3) require validated experimental assays~\cite{Richards2015ACMG}. Conflating these tiers risks miscalibrating clinical confidence.

    Closing this gap increasingly depends on scalable functional assays alongside improved annotation. Multiplexed assays of variant effect (MAVEs) and saturation genome editing can generate functional readouts for thousands of variants in parallel, while integrating these datasets with other lines of evidence is itself an emerging AI task~\cite{findlay2018accurate}. However, such assays are unavailable for many genes and variant classes, are gene- and context-specific, and their outputs need careful reconciliation with clinical phenotype data before they can inform variant classification. Heterogeneous evidence quality, inconsistent nomenclature, and continuously evolving biomedical knowledge further compound this challenge, making expert curation indispensable even where high-throughput functional data exist.

    Trustworthy clinical translation requires AI-assisted annotation and evidence synthesis to support prioritisation and curation, functional assays to establish biological effects, and expert interpretation to adjudicate the evidence, with AI complementing rather than substituting for experimental validation.

\subsection{Implications for Clinical Reliability}

    Variant interpretation uncertainty, limited evidence for rare variants, and incomplete functional validation constrain the reliability of AI-supported genomic decision-making. Conflicting findings and inconsistent annotation across databases can propagate through evidence integration into downstream predictions and clinical recommendations.

    In precision oncology, variant misclassification or overestimation of pathogenicity can directly affect therapeutic decisions and clinician confidence in AI-assisted systems. Addressing these risks requires stronger evidence generation, data curation, functional validation, and uncertainty-aware reporting alongside advances in NLP and machine learning. These limitations also affect how uncertainty can be estimated and communicated in clinical AI.

\section{Explainability and Uncertainty in Genomic Reasoning}\label{sec4}

\subsection{Model Interpretability}

    In cancer genomics, deep learning and large language models increasingly support variant interpretation and therapeutic decision-making by integrating genomic, clinical, and literature-derived evidence. Clinical use requires not only accurate predictions but also explanations that allow recommendations to be traced to their supporting literature, databases, and clinical observations. Argumentation-based frameworks extend this approach by providing structured justifications aligned with clinical reasoning~\cite{Ilgen2025PHAX}.

    Explanations are only as reliable as the evidence underlying model predictions. Interpretability should expose and communicate this evidence rather than compensate for weaknesses in the underlying knowledge base. Explainable artificial intelligence (XAI) methods aim to increase the transparency of complex models~\cite{Zhang2026ExplainableCancerAI}. Feature-attribution methods, including SHAP~\cite{NIPS2017_8a20a862}, Integrated Gradients~\cite{pmlr-v70-sundararajan17a}, saliency maps, and attention-based approaches, highlight input features associated with individual predictions. Counterfactual and example-based explanations provide complementary perspectives by showing how changes in input characteristics or similarities to previous cases relate to model outputs. In cancer genomics, these techniques have been explored for variant prioritisation, biomarker identification, and molecular subtype classification~\cite{10.1093/bib/bbad200}.

    Many explanation methods describe correlations learned by models rather than underlying biological mechanisms. Attention weights, for example, do not necessarily reflect causal reasoning~\cite{jain-wallace-2019-attention}, while feature-attribution scores may disagree across explanation methods despite identical model predictions~\cite{krishna2025disagreementproblemexplainablemachine}. These limitations are particularly relevant for foundation models and large language models, where convincing explanations may coexist with uncertainty or incomplete evidence.

   For clinical implementation, interpretability enables rather than guarantees trustworthy AI. Transparent explanations can support clinician understanding, verification of AI-generated recommendations, and regulatory accountability, but cannot compensate for unreliable evidence, poorly calibrated models, or insufficient clinical validation. What matters is whether explanations allow clinicians to critically evaluate model outputs against available biomedical evidence. Interpretability complements uncertainty communication and human expertise as part of trustworthy clinical reasoning.

\subsection{Uncertainty in Genomic Evidence}

    Uncertainty is an inherent characteristic of cancer genomics and represents one of the principal challenges for trustworthy AI-assisted clinical decision-making. Unlike uncertainty arising from model behaviour, genomic uncertainty originates from the underlying evidence itself. Cancer genomes are highly heterogeneous, many variants are insufficiently characterised, and the available biomedical literature frequently contains incomplete, conflicting, or evolving evidence. Consequently, genomic uncertainty is multifactorial, arising from incomplete biological knowledge, probabilistic evidence, biological ambiguity, heterogeneous data resources, and the complexity of integrating genomic and clinical information~\cite{Richards2015ACMG,Han2017Taxonomy}. AI systems do not reason over definitive biological knowledge but over evidence that is often probabilistic, fragmented, and continuously evolving.

    Genomic uncertainty arises from interconnected sources, including variants of uncertain significance (VUS), limited functional validation, inconsistent findings across studies, and population-specific differences in genomic datasets~\cite{Richards2015ACMG, Popejoy2016}. Rare clinically relevant variants also contribute to uncertainty because of limited experimental evidence and sparse representation in public databases. Although AI models can integrate scientific literature, genomic databases, and clinical data, they cannot eliminate uncertainty embedded within these sources and may propagate or amplify it when conflicting or incomplete evidence is inadequately represented. The reliability of AI outputs depends not only on model performance but also on the quality and certainty of the underlying evidence.

    The increasing use of large language models and automated evidence synthesis further highlights this challenge~\cite{Singhal2023}. Although these systems can integrate heterogeneous information into coherent clinical summaries, the confidence conveyed by their outputs may not reflect the quality or completeness of the underlying evidence~\cite{info:doi/10.2196/66917}. Fluent explanations may create an impression of certainty despite limited or conflicting genomic evidence, potentially leading clinicians to overestimate the reliability of AI-generated recommendations~\cite{Tang2023}. Distinguishing confidence in model predictions from confidence in the biomedical evidence supporting them is essential for responsible clinical interpretation.

    Addressing uncertainty in cancer genomics requires more than improvements in predictive performance. AI systems should explicitly communicate the strength, provenance, and limitations of the evidence supporting their recommendations, enabling clinicians to distinguish well-established genomic knowledge from preliminary or uncertain findings, and to interpret AI-generated recommendations within their appropriate evidential and clinical context.

\subsection{Model Uncertainty}

    While the previous section focused on uncertainty in genomic evidence, AI systems introduce additional uncertainty through the modelling process itself. Trustworthy clinical translation requires consideration of both evidence uncertainty, arising from limitations in the underlying biomedical knowledge, and model uncertainty, arising from the behaviour and assumptions of the AI system.

    Modern machine-learning models typically operate as probabilistic function approximators, making their predictions sensitive to the training-data distribution, model architecture, parameter estimation, and input variability. High predictive confidence does not necessarily imply correctness: miscalibrated models may assign excessive confidence to erroneous outputs~\cite{info:doi/10.2196/66917, pmlr-v240-bajwa24a}, while their uncertainty estimates may not accurately reflect the reliability of individual predictions. In clinical genomics, discrepancies between reported confidence and prediction reliability can affect variant classification, prioritisation, and treatment selection.

    These challenges are particularly pronounced in large language models and other foundation models increasingly used for evidence synthesis, clinical summarisation, and genomic interpretation. LLMs may produce statements that are linguistically plausible yet factually incorrect, include unsupported rationales, or convey excessive confidence despite limited or conflicting biomedical evidence~\cite{Singhal2023}. Coherent narratives and well-structured explanations may obscure underlying uncertainty, mask important caveats, and make erroneous recommendations appear credible. Addressing model uncertainty calls for calibration, uncertainty quantification, confidence estimation, and transparent communication of model limitations, alongside the evidence-level safeguards discussed above~\cite{Dubey2026UbiQTree, 10.1016/j.compbiomed.2023.107441}.

\subsection{Trustworthy Clinical Decision Support}

   In clinical genomics, AI systems increasingly function as decision-support tools rather than autonomous decision-makers, making their value dependent not only on predictive performance but also on how clinicians interpret and act on their outputs~\cite{Pierce2022Explainability}. Clinicians must decide when to accept, question, or override AI-assisted recommendations, often under time pressure and incomplete information. Poorly quantified or communicated uncertainty may lead to over-reliance on uncertain recommendations or under-use of potentially beneficial support. Appropriate reliance requires calibrated uncertainty estimates aligned with clinically meaningful decision thresholds, transparent communication of model limitations, and human--AI collaboration that supports rather than replaces clinical judgement~\cite{info:doi/10.2196/66917}.

    Communicating uncertainty at the point of care is challenging because even well-quantified uncertainty may be interpreted differently depending on how it is presented. AI-based clinical decision-support systems communicate uncertainty in different ways, from risk scores and categorical labels to visualisations and explanatory text~\cite{10.1145/3830235, Kompa2021SecondOpinion}. When uncertainty information is overly complex, inconsistent, or poorly integrated into clinical workflows, clinicians may ignore it and interpret model outputs simply as "correct" or "incorrect"~\cite{Abbas2025ExplainableAI}. Conversely, overly cautious or alarmist communication may erode trust by making systems appear unreliable or difficult to interpret.

    In genomic contexts, uncertainty must be communicated across different audiences and tasks, from laboratory specialists reviewing variant evidence to clinicians counselling patients and multidisciplinary tumour boards integrating multiple data sources. This calls for layered representations that combine concise, decision-oriented summaries with access to detailed evidence traces, model limitations, and context-specific caveats. Such designs can support both routine decisions and deeper examination of complex cases, helping clinicians align their reliance on AI recommendations with the strength and scope of the supporting evidence.

\section{Data Governance and Reproducibility Barriers}\label{sec5}

\subsection{Data Standardisation, Governance, and Reproducibility}
    Besides computational advances, the clinical translation of NLP-enabled AI in cancer genomics depends on the quality, governance, and reproducibility of the underlying data. Despite progress in literature mining, clinical text understanding, and evidence synthesis, these systems are constrained by biomedical data interoperability.

    Genomic and clinical information is distributed across clinical reports, pathology narratives, sequencing data, biomedical literature, and electronic health records. These sources use different formats, local terminologies, and ontologies, creating substantial heterogeneity. NLP can help harmonise these resources by mapping local terms to standard vocabularies, such as HGVS, SNOMED CT, and ICD-O, and aligning metadata. However, its performance and generalisability still depend on consistent annotation and high-quality reference data. Differences in reporting practices, patient populations, and annotation protocols can cause dataset shift and reduce model reliability across clinical settings.

    Standards and data models such as OMOP, HL7 FHIR, LOINC, UMLS, HGVS, and the FAIR principles provide a basis for harmonising data across institutions~\cite{VanDerVelde2022FAIRGenomes}. International initiatives, including the Global Alliance for Genomics and Health (GA4GH)~\cite{Rehm2021GA4GH}, Cancer Research Data Commons (CRDC)~\cite{10.1158/0008-5472.CAN-23-2655}, ICGC Accelerating Research in Genomic Oncology (ICGC-ARGO)~\cite{NahalBose2025ICGCARGO}, and Beyond 1 Million Genomes~\cite{Saunders2019MillionGenomes}, extend these efforts to large-scale data sharing. Adoption remains uneven across healthcare systems, however, and this unevenness, rather than a lack of standards, continues to limit reproducibility in practice~\cite{Belien2022FAIRGenomes}.

    Governance is a complementary requirement: beyond protecting patient privacy, it establishes mechanisms for data access, provenance, accountability, and quality assurance across the AI lifecycle. The European Health Data Space (EHDS) provides a regulatory framework for the secondary use of electronic health data, including genomic and multi-omics data~\cite{EHDSArt51}, and supports their use in training, testing, and evaluating AI systems~\cite{EHDSArt53}. It also mandates secure data access environments and introduces standardised data quality and utility labels covering completeness, consistency, representativeness, and quality-management procedures~\cite{EHDSArt73, EHDSArt78}. These labels provide a shared basis for assessing and communicating dataset fitness. Alongside the FAIR principles and the AI Act, the EHDS positions governance not as a compliance afterthought but as part of the infrastructure for reliable clinical AI.

\subsection{Emerging AI Systems and Governability Challenges}
    Recent advances in foundation models, large language models (LLMs), and multimodal AI have expanded AI capabilities in cancer genomics~\cite{Tsang2025FoundationModels}. Unlike task-specific NLP systems, these models are increasingly being developed to integrate genomic data, biomedical literature, clinical records, pathology reports, imaging, and molecular databases, extending AI applications from information extraction to evidence synthesis, variant interpretation, and hypothesis generation~\cite{Menon2026FoundationEmbeddings}.

    This expansion introduces governance challenges beyond those of task-specific systems. Foundation-model behaviour depends not only on model architecture but also on evolving training data, retrieval resources, and model updates. A clinically deployed LLM updated without formal revalidation may, for example, drift from the guidelines against which it was originally validated. Reproducibility, traceability, and independent validation become harder to establish. Rapid changes in biomedical evidence add another challenge: a model trained on an earlier evidence base may still recommend treatments that have since been superseded. Trustworthy deployment requires continuous knowledge updating and revalidation rather than one-time validation.

    Proprietary deployment further complicates governance. Limited access to training data, model parameters, update histories, and system documentation constrains independent auditing. Even models with strong benchmark performance may leave healthcare institutions unable to verify the basis of a recommendation or determine whether model behaviour has changed since the last evaluation.

    These shifts move the focus of governance from evaluating individual predictions to managing the lifecycle of continuously evolving systems~\cite{ElArab2026LifecycleGovernance}, including version control, provenance tracking, auditability, and post-deployment monitoring across institutions and populations. The European AI Act's emphasis on post-market monitoring for high-risk systems reflects this shift~\cite{AIAArt72} and implies that pre-deployment validation alone is insufficient. Sustained reliability depends on ongoing evaluation, record-keeping, and post-deployment oversight throughout the system lifecycle.

\subsection{Privacy, Bias, and Data Stewardship}

    The standardisation and governance challenges described above concern how genomic and clinical data are structured and shared; a related challenge is who may access these data and under what safeguards. Cancer genomics involves sensitive information because it links identifiable genetic, clinical, and phenotypic data. NLP systems operating on EHRs, pathology reports, and genomic records depend on robust safeguards, including de-identification of textual and genomic data, differential privacy techniques, and secure processing environments with strong access controls. Combining multiple data modalities may further increase re-identification risk beyond that posed by individual data sources.

    These concerns are addressed by legal frameworks such as the EU GDPR and US HIPAA, which regulate the handling of personal and protected health information within their respective scopes; the GDPR additionally classifies genetic data as a special category subject to stricter safeguards. In practice, data use requires an appropriate lawful basis and adherence to principles such as data minimisation and purpose limitation. Beyond HIPAA's federal requirements, genetic data protections in the US may vary across states, with some state laws providing additional rights concerning genetic information~\cite{Gitter2023GeneticPrivacy}. Table~\ref{tab:gdpr-hipaa} compares the two frameworks across consent, erasure, breach notification, and individual data rights, highlighting differences relevant to multi-institutional cancer genomics research involving EU and US sites.

\begin{table}[h!]

\caption{Comparison of GDPR and HIPAA provisions relevant to genomic and clinical data governance.}
\small

\begin{tabular}{p{3.5cm} p{4.25cm} p{4.25cm}}
\hline
\textbf{Feature} & \textbf{GDPR (EU)} & \textbf{HIPAA (US)} \\
\hline
    Scope & Applies broadly to personal data; genetic and health data are classified as special categories subject to additional protections & Applies to Protected Health Information (PHI) held or transmitted by covered entities and their business associates \\
\hline
    Consent / Authorisation & Processing must have a lawful basis and, for genetic or health data, an applicable condition for processing special-category data; consent is one possible basis & Authorization is generally not required for treatment, payment, and healthcare operations, but is required for many other uses and disclosures \\
\hline
    Erasure & Individuals may request erasure under specified conditions, subject to exceptions & No equivalent general right to require erasure of PHI \\
\hline
    Breach Notification & Supervisory authorities must generally be notified within 72 hours where notification is required; individuals must be informed when the breach is likely to result in high risk & Affected individuals must generally be notified without unreasonable delay and no later than 60 days; additional notification requirements apply to larger breaches \\
\hline
    Individual Data Rights & Provides rights including access, rectification, erasure, restriction, and data portability, subject to applicable conditions and exceptions & Provides rights including access to and amendment of PHI and an accounting of certain disclosures \\
\hline
\end{tabular}

\vspace{2pt}
{\footnotesize \textit{Sources: EU General Data Protection Regulation (GDPR) and the US HIPAA Privacy and Breach Notification Rules.}}

\label{tab:gdpr-hipaa}
\end{table}

    Questions about who controls genomic data and how these data can be used are not fully resolved. Governance also involves collective interests, including community involvement and indigenous data sovereignty~\cite{carroll2020care}. Restrictions on data reuse may create tensions between data sovereignty and cross-institutional standardisation and reproducibility. A related governance challenge is bias in non-representative training data, which can contribute to healthcare disparities~\cite{obermeyer2019dissecting, rajkomar2018ensuring}. In precision oncology, this problem is compounded by rare-variant evidence scarcity: populations underrepresented in genomic databases may also have less supporting evidence available for NLP-based systems, reinforcing biases in both the evidence base and model training. Mitigation calls for diverse training data, bias audits, and fairness-aware methods, but algorithmic approaches cannot substitute for representative data. Privacy-preserving approaches such as federated learning may help broaden data representation without centralizing sensitive information.

    Trustworthy use of genomic data also depends on patients understanding how their data are used in AI-enabled care. Consent should be explicit, contextual, and revocable, with clear communication of risks, benefits, and the role of AI and NLP in genomic testing. Meaningful consent also requires transparency about the supporting evidence and clear mechanisms for human oversight, audit, and redress.

\section{Interoperability and Multimodal Fragmentation}\label{sec6}

    NLP supports multimodal cancer genomics by transforming clinical narratives and biomedical literature into structured representations that can be integrated with genomic, imaging, and molecular data~\cite{Yang2024MultimodalDL}. Foundation models and multimodal AI further support the integration of heterogeneous information for disease modelling, biomarker discovery, and variant interpretation~\cite{Truhn2024LLMOncology, Li2026FoundationModels}. However, multimodal integration is a major barrier to clinical translation~\cite{Sweeney2023BigDataCancer}. NLP can reduce semantic fragmentation by harmonising unstructured biomedical information, but cannot resolve fragmentation caused by heterogeneous data infrastructures, incompatible acquisition pipelines, incomplete multimodal records, or the lack of agreed frameworks for cross-institutional data exchange.

    Discussions of interoperability in genomics tend to collapse these distinct obstacles into a single technical problem. The European Interoperability Framework (EIF) offers a more discriminating vocabulary~\cite{EC2017EIF}. Developed to support cross-border digital public services, and given partial binding force for the public sector by the Interoperable Europe Act~\cite{InteroperableEuropeAct2024}, the EIF separates interoperability into four layers, namely legal, organisational, semantic, and technical, underpinned by cross-cutting interoperability governance (Table~\ref{tab:eif-layers}). Its target setting is structurally similar to multi-institutional cancer genomics, in which many autonomous organisations, operating under different jurisdictions and mandates, attempt to combine data they did not jointly design; and its layers form a dependency chain, read top-down, whose asymmetry is discussed below.

    Read through this lens, two claims organise the remainder of this section. The first is that the field's interoperability effort is unevenly distributed. Cancer genomics has invested heavily in the two lower layers and comparatively little in the two upper ones, yet it is the upper layers that most often determine whether a multi-institutional study proceeds at all. The second, more consequential for trustworthiness, is that interoperability failures are characteristically \emph{silent}. A missing file or malformed message fails loudly and is therefore fixed; a many-to-one concept collapse, a unit mismatch, a staging-version discrepancy, or a well-formed but incorrect terminology code does not, but passes schema validation and propagates into a confidently calibrated output carrying no indication of the defect upstream. An entire class of interoperability failure is thus undetectable at the point where it does damage, and trustworthiness requires above all that failures be detectable.

%---------------- TABLE 2 ----------------

\begin{table}[t]
  \centering
  \small
  \caption{European Interoperability Framework (EIF) applied to multi-institutional cancer genomics, showing its four interoperability layers and cross-cutting governance, with instruments adapted to the clinical genomics setting~\cite{EC2017EIF}.}
  \label{tab:eif-layers}
  \begin{tabularx}{\textwidth}{@{}
    l
    >{\raggedright\arraybackslash}p{3.3cm}
    >{\arraybackslash}X
@{}}
    \toprule
    \textbf{Layer} & \textbf{Core question} &
    \textbf{Adapted instruments} \\
    \midrule
    Legal &
      Is the exchange lawful and legally valid? & EHDS data permits and secure processing environments; MDR/IVDR conformity assessment; AI Act high-risk obligations; GDPR Art.~9 and national research derogations \\
      \addlinespace
    Organisational &
      Who does what, under what commitment? & Consortium and data-sharing agreements; curation and stewardship roles; multi-centre validation protocols; molecular tumour board procedures\\
    \addlinespace
    Semantic &
      Is the data understood as intended? & Standard terminologies (HGVS, SNOMED CT); clinical significance criteria (ACMG/AMP); curated knowledge bases (ClinVar, CIViC) \\
    \addlinespace
    Technical &
      Can systems connect and exchange data? & Common data models (HL7 FHIR, OMOP); sequence and imaging formats; federated query and retrieval interfaces (GA4GH) \\
    \midrule
    \rowcolor{gray!12}
    \multicolumn{3}{@{}p{\textwidth}@{}}{\footnotesize
      \textbf{Cross-cutting governance:} arrangements for agreeing, maintaining, and versioning these instruments across institutions, alongside clinical governance
and trustworthy AI assessment.} \\
    
    \bottomrule
  \end{tabularx}
\end{table}

\subsection{Semantic Interoperability as Lossy Compression}

    Most standardisation efforts in cancer genomics address the technical and semantic layers discussed in Section~\ref{sec5}. Technical-layer obstacles are real but comparatively tractable, and they fail loudly: a tumour sequenced against one reference genome build and imaged with a pathology scanner using institution-specific colour calibration cannot be pooled with data from another site, and a pipeline encountering such a mismatch typically stops. The semantic layer is where the harder and quieter problems accumulate.

    Semantic harmonisation is inherently lossy. Mapping local codes to shared terminologies is often many-to-one and non-invertible, so clinically meaningful distinctions such as laterality, qualifiers, or local subtypes may be discarded to enable cross-site comparability. These mappings trade local fidelity for interoperability, yet measuring the magnitude and subgroup distribution of the resulting information loss is an important unmet need. Models trained on harmonised data cannot recover distinctions that have already been removed from their inputs, nor can they signal that such distinctions ever existed. Standards proliferation compounds the problem: datasets may be represented in FHIR, OMOP, Phenopackets, mCODE, or local models, requiring additional mappings between standards that introduce further loss and are rarely validated end to end.

    Meaning is also versioned, and genomic data differ from the administrative setting of the EIF in how revisions affect previously issued records. Diagnostic terminologies, disease coding systems, staging systems, response criteria, and reference genome builds are revised on different schedules, while the underlying evidence base evolves independently (Section~\ref{sec3}). The same variant call file can carry a different clinical meaning a year later without the file itself changing. Mapping between terminology versions cannot recover this difference because the evidence, rather than the vocabulary, has changed. Provenance, source version, and classification date are part of the meaning being transferred, not merely metadata; harmonising terminology without this context can align vocabulary while allowing meaning to drift.

    The existence of a standard is not sufficient. Tumour mutational burden (TMB), for example, can vary for the same tumour depending on panel composition, gene coverage, and bioinformatics pipeline, making it partly assay-dependent~\cite{Merino2020TMB}. Radiomic features are similarly sensitive to acquisition and processing, motivating initiatives such as the Image Biomarker Standardisation Initiative~\cite{Zwanenburg2020IBSI}. Derived variables may appear equivalent across sites while remaining substantively different. The same problem applies to clinical endpoints such as index date, real-world progression-free survival, and time to next treatment, which can be syntactically harmonised without being clinically comparable. Linguistic disparities add another layer: clinical NLP is concentrated in well-resourced languages, leaving harmonisation less reliable for languages with limited NLP resources, a gap that goes unrecorded in the harmonised data itself.

\subsection{Organisational and Legal Interoperability}

    The organisational layer concerns who does what and under what commitment: which institution re-curates a shared variant knowledge base, who is accountable when reclassification invalidates an issued clinical report, and who re-runs validation when a partner site changes its pipeline. Prospective multi-centre validation is an organisational commitment before it is a technical exercise, helping explain its scarcity.

    Organisational constraints are often mistaken for technical ones. One example is mapping rot: mappings created as project deliverables may become progressively inaccurate as terminologies evolve if they are not maintained, while resulting errors may remain undetected by downstream users. Sustained harmonisation requires recurring operational work, often performed by data-holding institutions with limited dedicated funding or academic credit. Expertise is another constraint, as staff fluent in both clinical oncology and relevant data models are scarce, while negotiated access and permitting may be too slow for models requiring frequent updates. More fundamentally, health data architectures often reflect the organisational structure of the health systems that produced them, limiting what technical harmonisation alone can achieve.

    Federated approaches are often presented as a solution to cross-border data sharing~\cite{10.1007/978-3-031-08999-2_1}, and are now being deployed at scale through initiatives such as UNCAN-CONNECT~\cite{UNCANCONNECT2025}. Yet participation requires trusted research environments, data stewardship, harmonised mappings, and, for imaging or foundation-model workloads, local accelerated compute. These requirements favour large, urban, well-resourced cancer centres whose populations may differ systematically from regional and community settings. Federation thus democratises \emph{analysis}, not \emph{participation}: compute asymmetry determines which nodes contribute, reintroducing the representativeness bias described in Section~\ref{sec5}.

    The legal layer concerns whether cross-border or inter-institutional data exchange is lawful. The main challenge is not GDPR’s classification of genetic data as sensitive~\cite{gdpr2016}, but the national implementation of its research derogation, which creates divergent permissions across jurisdictions. Thus, identically formatted data may be subject to different access conditions, making aggregate- and discovery-level federation easier than record-level access. The EHDS addresses this layer through health data access bodies and data permits intended to support secondary use on a common basis rather than through case-by-case negotiation~\cite{EHDSArt57,EHDSArt68}.

    The AI Act makes harmonisation a legal rather than merely scientific requirement. For high-risk systems, training, validation, and test data must be relevant, sufficiently representative, and, as far as possible, error-free and complete for their intended purpose~\cite{AIAArt10, AIAArt8to15}. In clinical genomics, meeting and documenting these requirements depends on the semantic harmonisation described above and must be demonstrated in conformity assessment.  An unresolved issue is whether mapping and transformation layers fall within the regulated boundary. Because changes to these layers can alter a clinical decision-support tool’s behaviour, they may constitute design changes requiring revalidation under the medical-device framework~\cite{MDRArt2, MDRRule11, IVDR2017}. Yet such pipelines are often maintained as research infrastructure without downstream revalidation, making their regulatory status consequential for the cost of continuous harmonisation.

    The dependency between layers is asymmetric: legal failure disables the layers below, while technical excellence cannot compensate for it. A fully harmonised FHIR-based exchange cannot proceed without a lawful basis and permit, whereas permitted institutions can exchange data through less sophisticated technical means if necessary. Investment in lower layers yields returns only once upper-layer constraints are resolved, which may help explain why standards adoption in genomics has outpaced cross-institutional data flow~\cite{Belien2022FAIRGenomes}.

\subsection{When the Harmonisation Layer Is Itself an AI System}

    Language models increasingly perform semantic harmonisation tasks formerly handled by human terminologists, including local-code mapping, concept extraction from narrative text, and vocabulary reconciliation. These applications of the capabilities surveyed in Section~\ref{sec2} are often the only tractable means of harmonising at scale for many institutions. Yet they introduce a consequence largely overlooked in the responsible-AI literature: non-deterministic, versioned, and often proprietary models enter the clinical evidence chain \emph{upstream} of the systems to which oversight is actually applied.

    Three problems follow. First, model-assisted harmonisation creates reproducibility debt: a dataset cannot be regenerated without the exact model version, and proprietary models subject to silent revision may make reproduction impossible in principle, despite reproducibility being harmonisation’s purpose (Section~\ref{sec5}). Second, its characteristic error is well formed: a model can return a syntactically valid but incorrect mapping that downstream systems cannot distinguish from a correct one.

    The third is particularly consequential for fairness (Section~\ref{sec5}), as these errors are not randomly distributed. A language model maps common, well-documented, English-language oncology concepts more reliably than rare tumour types, paediatric entities, minority-language narratives, and legacy local vocabularies, so harmonisation error rates correlate precisely with the subgroups fairness auditing exists to protect. Because these errors enter the dataset as ground truth, they are upstream of, and invisible to, every fairness audit performed on models trained from it; bias introduced at the harmonisation layer cannot be detected by evaluating the downstream model, which is where essentially all current auditing effort is directed.

    This does not argue against model-assisted harmonisation, often the only viable option. It argues for treating harmonisation as a clinical AI component in its own right. Mapping performance should be established against gold-standard benchmarks, with human inter-annotator agreement also reported as a reference. Mappings affecting clinical data should be human-reviewed with the reviewer recorded and each should carry provenance covering model identity, version, configuration, and confidence. Most importantly, the system should be required to abstain rather than guess. An explicit refusal to map converts a silent error into a loud one, making the failure detectable before it propagates downstream.

    Structured trustworthy-AI assessment can address these requirements if its scope includes the harmonisation layer. Approaches such as Z-Inspection use interdisciplinary socio-technical scenarios and can be applied post-hoc or alongside development~\cite{Zicari2021ZInspection}. This distinction matters more than timing: post-hoc assessment of the deployed model is blind to upstream harmonisation errors for the same reason downstream fairness audits are, whereas co-design can examine mapping decisions, abstention behaviour, and provenance as they develop. Neither approach reaches harmonisation automatically; this depends on where the socio-technical scenario boundary is drawn, which current practice almost invariably places around the clinical tool rather than the pipeline supplying it.

\subsection{Multimodal Foundation Models and the Fragmented Data Ecosystem}

    Individual modalities are often incomplete or collected at different time points: patients with rich genomic data may lack corresponding imaging or pathology records, and vice versa. This asynchrony, rather than any single missing standard, complicates multimodal learning and external validation in practice.

    Large multimodal models may appear to bypass the semantic and technical layers of Table~\ref{tab:eif-layers} by learning directly from heterogeneous inputs. Yet statistically absorbing heterogeneity does not remove it: a visible failure, such as an unmappable code or failed data join, may instead become invisible when a model produces a confident output from inputs whose provenance and comparability were never established. Nor does model capability address the upper two layers: no foundation model provides a lawful basis for combining cohorts or assigns responsibility for revalidating shared resources as their evidence base changes.

    Asynchrony should also be distinguished from interoperability because it can persist within a single, well-governed institution in which all four layers are already satisfied. Missingness in multimodal oncology data may reflect clinical trajectory rather than protocol, for example when sequencing occurs late in treatment or imaging only after progression. Models may learn aspects of that trajectory alongside the underlying biology. Addressing this requires prospective, protocol-driven multimodal collection rather than further harmonisation of existing data.

\subsection{Barrier Mitigation}

    These barriers compound in multi-institutional settings~\cite{Conway2019NGSWorkflow}. Models trained within one institutional data ecosystem may degrade when deployed elsewhere because of differences in technical and procedural context rather than underlying biology.

    The silent-failure argument exposes a tension often overlooked in the literature: federated architectures favour privacy by default, but hidden distributions cannot be inspected for batch effects, label skew, or confounding, and fairness cannot be verified for uncollected attributes. Federation does not remove these hazards; it removes the view of them. Verification should therefore be a primary service of federated infrastructure: distribution diagnostics, site-stratified reporting, and per-node performance deltas can restore inspection without moving records, providing the visibility required for trustworthy-AI assessment. \emph{Federated external validation} may consequently be more valuable than federated training, since evaluating models against locally held data and reporting site-level differences is cheaper, more immediately deliverable, and more directly addresses generalisation failures than distributed model fitting. More broadly, interoperability requires coordinated legal, organisational, semantic, and technical commitments, supported by governance across all four as developed in Sections~\ref{sec7} and~\ref{sec8}; the principal limitation is no longer the availability of capable AI models, but the fragmented clinical data ecosystem required for reliable deployment.

\section{Clinical Implementation of High-Risk AI}\label{sec7}
\subsection{Regulatory Requirements for High-Risk AI}
   
    As NLP-enabled AI systems increasingly support genomic interpretation, clinical decision support, and precision oncology, they fall within regulatory frameworks governing high-risk AI. Under the European AI Act, an AI system that is itself a product, or a safety component of a product, regulated under the Medical Device Regulation (MDR) or In Vitro Diagnostic Medical Devices Regulation (IVDR) is classified as high-risk where both conditions of Article~6(1) are met: the AI system is intended to be used as a safety component of such a product, or is itself such a product, and the product is required to undergo third-party conformity assessment under the Union harmonisation legislation listed in Annex~I~\cite{AIAArt6AnnexI, MDRArt2, IVDR2017, MDRRule11}. Clinical genomics AI systems meeting these conditions are subject to requirements extending beyond predictive performance to safe deployment, transparency, human oversight, and lifecycle monitoring.

    Compliance consequently extends beyond model validation. Providers must implement risk management, technical documentation, data governance, and measures ensuring robustness, accuracy, and cybersecurity~\cite{AIAArt8to15}. These requirements may also be relevant where general-purpose AI (GPAI) models are integrated into clinical applications, while providers and deployers of the resulting high-risk AI systems are subject to distinct obligations~\cite{AIAArt16,AIAArt26}.

    Reuse creates an additional documentation burden because providers of GPAI models must supply sufficient information and documentation to downstream providers integrating those models into AI systems~\cite{AIAArt53}. Although qualifying free and open-source GPAI models are exempted from certain technical-documentation and downstream-information duties~\cite{AIAArt53_2}, this exemption does not extend to GPAI models with systemic risk~\cite{AIAArt51,AIAArt55}. Compliance becomes shared, ongoing accountability across the model lifecycle rather than a single pre-market assessment.

    These obligations are not merely external constraints on innovation. A clinician who cannot determine why a model flagged a variant, or whether it has been updated since relevant guidance changed, has little basis for trusting its output regardless of benchmark accuracy, precisely the gap lifecycle regulation is intended to address.

\subsection{Human Oversight and Accountability}

    Clinical genomics is a domain in which AI should support, not replace, human expertise. Variant interpretation, diagnosis, treatment selection, and patient management often depend on uncertain, incomplete, or evolving evidence requiring clinical judgement beyond computational prediction. AI recommendations should be critically evaluated by healthcare professionals rather than treated as autonomous clinical decisions.

    Meaningful oversight requires more than reviewing outputs. Clinicians must be able to understand a recommendation's rationale, recognise unreliable predictions, and challenge or override AI-assisted decisions. This relies on transparent explanations, communicated uncertainty, evidence provenance, and workflows that enable meaningful clinician--AI interaction and independent clinical verification.

    Accountability rests with healthcare professionals and the institutions deploying these systems. Governance should define responsibilities across developers, providers, and clinical users throughout the AI lifecycle, while documentation, audit trails, and traceability enable retrospective assessment of AI-assisted decisions. External audits and clearly defined redress pathways extend accountability beyond internal institutional oversight.

\subsection{Post-Deployment Monitoring}

    Clinical validation before deployment is necessary but insufficient for AI systems in cancer genomics, where scientific knowledge, clinical guidelines, genomic databases, and population characteristics evolve over time. Post-deployment monitoring should assess real-world performance, detect dataset shift and degradation, document adverse events, and reassess calibration as new variants, functional evidence, and therapeutic guidance emerge. Version control, documentation, provenance tracking, and periodic revalidation are likewise needed to identify changes in system behaviour and preserve reproducibility, traceability, and regulatory compliance, particularly for foundation models that are updated or draw on external knowledge sources. Trustworthy deployment should thus be treated as continuous governance rather than a one-time regulatory milestone~\cite{AIAArt72, dudder2021ethical}, with sustained monitoring and maintenance needed to preserve clinician confidence, patient safety, and long-term reliability. Table~\ref{tab:lifecycle} summarizes these requirements across the full AI system lifecycle, from development through maintenance.

\begin{table}[h]
\centering
\caption{Clinical implementation requirements across the AI system lifecycle.}
\label{tab:lifecycle}

\renewcommand{\arraystretch}{1.45}
\setlength{\tabcolsep}{8pt}

\begin{tabularx}{0.90\linewidth}{
>{\raggedright\arraybackslash}p{3.2cm}
>{\raggedright\arraybackslash}X
}

\toprule
\rowcolor{gray!15}
\textbf{Lifecycle stage} &
\textbf{Core implementation requirement} \\
\midrule

\rowcolor{teal!5}
\textbf{01 Development} &
Governance-by-design, data quality assurance, risk management, and technical documentation. \\

\rowcolor{teal!9}
\textbf{02 Validation} &
External and multi-institutional validation, calibration assessment, and transparent reporting of limitations. \\

\rowcolor{teal!13}
\textbf{03 Deployment} &
Meaningful human oversight, workflow integration, traceability, and clearly defined clinical responsibilities. \\

\rowcolor{teal!17}
\textbf{04 Monitoring} &
Post-deployment surveillance, real-world performance evaluation, adverse-event reporting, and dataset-shift detection. \\

\rowcolor{teal!21}
\textbf{05 Maintenance} &
Version control, provenance tracking, periodic revalidation, and controlled updating of models and knowledge resources. \\

\bottomrule
\end{tabularx}
\end{table}

\section{Critical View and Pathways to Trustworthy Clinical Translation}\label{sec8}

\subsection{From Failure Domains to Integration Pathways}

    Rather than examining individual AI applications or implementation challenges in isolation, our critical view frames clinical translation as a systems-level problem arising from interactions among evidence inconsistency, explainability and uncertainty, data governance and reproducibility, and interoperability. Weaknesses across these domains can propagate through the translational pathway into downstream clinical decisions, limiting adoption and trust.

    Table~\ref{tab:4a_system_coverage} maps six representative systems onto the four translational failure domains. It is illustrative rather than exhaustive or comparative: each system serves a specific priority rather than a complete clinical workflow. ClinVar focuses on variant classification; LitVar on literature--variant linkage; PubTator on entity annotation rather than evidence synthesis or conflict resolution; CIViC/CIViCmine on curated clinical interpretation, with scalability constrained by manual curation; MatchMiner on trial matching; and DGIdb on drug--gene interaction aggregation without systematic conflict resolution. Their limitations therefore largely reflect intended scope rather than implementation deficiencies. Yet collectively, these systems tend to cover only one or two domains, and none addresses all four, indicating that translational failure is a structural ecosystem-level gap rather than the failure of a single tool. Table~\ref{tab:4b_failure_domains} summarizes the major translational failure domains, their clinical consequences, and the corresponding guiding principles for trustworthy clinical translation.
    
\begin{table*}[!htbp]
\centering
\caption{Translational failure domains in cancer genomics: system coverage and clinical implications.}
\label{tab:4_combined}

\begin{subtable}{\textwidth}
\centering
\caption{Coverage of representative systems supporting cancer genomics interpretation and clinical translation across the four translational failure domains. Markers reflect each system's stated scope and design focus as described in the cited literature rather than an empirical benchmarking exercise: \checkmark\ indicates direct coverage, $\triangle$ indicates partial or indirect coverage, and $\times$ indicates that the domain falls outside the system's stated scope.}
\label{tab:4a_system_coverage}
\resizebox{\textwidth}{!}{%
\footnotesize
\renewcommand{\arraystretch}{1.15}
\rowcolors{2}{gray!8}{white}
\begin{tabularx}{\textwidth}{@{}
  >{\raggedright\arraybackslash}X
  >{\centering\arraybackslash}X
  >{\centering\arraybackslash}X
  >{\centering\arraybackslash}X
  >{\centering\arraybackslash}X@{}}
\toprule
\rowcolor{gray!15}
\textbf{System} & \textbf{Evidence Inconsistency} & \textbf{Explainability \& Uncertainty} & \textbf{Data Gov. \& Reproducibility} & \textbf{Interoperability} \\
\midrule
    ClinVar~\cite{10.1093/nar/gkv1222} & \checkmark & $\times$ & $\triangle$ & $\triangle$  \\
    \addlinespace
    LitVar / \newline LitVar KG~\cite{10.1093/nar/gky355} & \checkmark & $\times$ & $\triangle$ & $\triangle$ \\
    \addlinespace
    PubTator~\cite{10.1093/nar/gkae235} & $\triangle$ & $\times$ & $\times$ & $\triangle$ \\
    \addlinespace
    CIViC / CIViCmine~\cite{Lever2019CIViC} & \checkmark & $\triangle$ & $\triangle$ & $\triangle$ \\
    \addlinespace
    MatchMiner~\cite{Klein2022MatchMiner} & $\times$ & $\times$ & $\times$ & \checkmark \\
    \addlinespace
    DGIdb~\cite{Griffith2013DGIdb} & $\triangle$ & $\times$ & $\times$ & $\triangle$ \\
\bottomrule
\end{tabularx}
}
\end{subtable}

\vspace{1.2em}

\begin{subtable}{\textwidth}
\centering
\caption{Major translational failure domains in cancer genomics, their clinical consequences, and corresponding guiding principles for trustworthy clinical translation.}
\label{tab:4b_failure_domains}
\resizebox{\textwidth}{!}{%
\renewcommand{\arraystretch}{1.35}
\setlength{\tabcolsep}{6pt}
\footnotesize
\rowcolors{2}{gray!8}{white}
\begin{tabularx}{\textwidth}{@{}p{0.19\textwidth} p{0.24\textwidth} p{0.26\textwidth} X@{}}
\toprule
\rowcolor{gray!20}
\textbf{Failure Domain} &
\textbf{Primary Challenge} &
\textbf{Clinical Consequence} &
\textbf{Guiding Principle (Box 1)} \\
\midrule
\textbf{Evidence \newline inconsistency} \newline \footnotesize(Sec.~\ref{sec3})&
    Conflicting, incomplete, or insufficiently validated genomic evidence &
    Variable variant interpretation and reduced confidence in clinical recommendations &
    Prioritize evidence quality over model complexity; integrate functional validation \\
\textbf{Explainability \& uncertainty} \newline \footnotesize(Sec.~\ref{sec4}) &
    Limited interpretability and poorly communicated uncertainty &
    Reduced clinician trust and difficult verification of AI-assisted decisions &
    Promote explainable and uncertainty-aware AI \\
\textbf{Data governance \& reproducibility} \newline \footnotesize(Sec.~\ref{sec5}) &
    Fragmented standards, inconsistent annotation, and limited provenance &
    Poor reproducibility, constrained data sharing, and limited external validation &
    Promote reproducibility and transparent evaluation; embed governance throughout the AI lifecycle \\
\textbf{Interoperability \& multimodal fragmentation} \newline \footnotesize(Sec.~\ref{sec6}) &
    Heterogeneous data infrastructures and incompatible clinical standards &
    Limited cross-institutional generalizability and unreliable clinical deployment &
    Strengthen interoperability across data ecosystems \\
\bottomrule
\end{tabularx}
}
\end{subtable}

\end{table*}

    The principles in Box~1 are interconnected: evidence quality, explainability, reproducibility, interoperability, governance, and human oversight reinforce one another across the translational pathway. Responsible integration requires a systems-level approach that addresses scientific, technical, clinical, and governance challenges together.

\clearpage
\begin{tcolorbox}[title=Box 1. Principles for Trustworthy Clinical Translation in Cancer Genomics,
    colback=gray!5,
    colframe=gray!40,
    colbacktitle=gray!15,
    coltitle=black]
\label{box1}
\small
\textbf{1. Prioritize evidence quality over model complexity}

Advanced AI models cannot compensate for incomplete, conflicting, or poorly validated genomic evidence.

\smallskip

\textbf{2. Promote explainable and uncertainty-aware AI}

Variant classifications and AI-generated recommendations should be accompanied by transparent explanations, evidence provenance, and calibrated uncertainty estimates.

\smallskip

\textbf{3. Integrate functional validation into AI-supported interpretation}

Computational predictions should be complemented by experimental evidence whenever possible.

\smallskip

\textbf{4. Promote reproducibility and transparent evaluation}

Models should be evaluated across diverse datasets and institutions with clear reporting of limitations and validation procedures.

\medskip

\textbf{5. Strengthen interoperability across data ecosystems}

Clinical adoption requires seamless integration of genomic, clinical, and biomedical knowledge sources.

\smallskip

\textbf{6. Embed governance throughout the AI lifecycle}

Privacy, accountability, transparency, and regulatory compliance should be incorporated from development through deployment.

\smallskip

\textbf{7. Maintain meaningful human oversight}

AI systems should support rather than replace expert clinical judgment.

\smallskip

\textbf{8. Adopt a risk-based approach to clinical deployment}

Oversight and validation requirements should be proportional to the potential clinical consequences of AI-assisted decisions.

\end{tcolorbox}

\subsection{Scaling Trustworthy Clinical Translation}

    Whereas Box~1 summarizes principles for individual AI systems, implementation also requires structural priorities beyond any single system or institution. Four stand out. First, cross-institutional and multilingual data infrastructures are needed for reproducibility and interoperability. Standards such as OMOP and HL7 FHIR, together with initiatives such as the EHDS, provide a foundation, but their value depends on adoption beyond well-resourced institutions and languages. Second, explainable and regulation-ready AI design should be treated as a starting requirement rather than a retrofit. Interpretability added only after model development may be insufficient for clinical or regulatory scrutiny; systems intended for high-risk clinical use are better served by explainability and audit mechanisms designed in from the outset. Third, closer integration of computational methods with experimental biology is needed to address the functional validation gap. Scaling multiplexed assays of variant effect and other high-throughput functional approaches, and integrating their outputs into AI-assisted interpretation, could reduce reliance on literature-derived evidence alone. Fourth, ethics, governance, and patient engagement must extend beyond compliance to questions of data control, community data sovereignty, and equitable access to AI-enabled precision oncology across low-resource and underserved health systems. Together, these priorities in infrastructure, explainable design, functional validation, and equitable governance provide a roadmap for trustworthy clinical implementation. Progress in one cannot fully compensate for weaknesses in the others: infrastructure without explainable design simply moves opaque systems into clinics faster, while governance without functional validation cannot correct evidence that was never reliable.

\section{Conclusion}\label{sec9}

    This review set out to examine how NLP and AI are used across the cancer genomics pipeline, which failure domains limit their trustworthy clinical translation, and which pathways may enable responsible clinical integration. The evidence reviewed here suggests that the field has largely resolved the first question: NLP-enabled AI systems now support literature mining, variant interpretation, multimodal integration, knowledge graph construction, and clinical trial matching at a scale unattainable through manual curation alone. It is the second and third questions, why these systems struggle to reach routine clinical practice and how that translational gap can be closed, that will determine the pace and impact of AI adoption over the coming decade. Closing this gap will require close collaboration among computational scientists, clinicians, molecular biologists, regulatory authorities, healthcare institutions, and patients.

    Building on the collaborative work of the EU Health Policy Platform Thematic Network on NLP for Cancer Genomics~\cite{eu2024jointstatement}, we hope this review provides a common conceptual framework for guiding future research, clinical implementation, and policy development across institutions and healthcare systems. The next generation of AI systems in cancer genomics will not be judged by predictive accuracy alone, but by their ability to generate clinically trustworthy evidence that can be interpreted, validated, governed, and sustained throughout routine oncology practice.

\section{Data Availability}
No new data were generated or analysed in support of this research.

\section{Author Contributions}
  B.I. and G.H. conceived the review. B.I. led the manuscript’s conceptual restructuring and drafted the manuscript. G.H., Y.T., D.K., P.P., D.H., K.L., and M.W. provided domain expertise and critically reviewed the manuscript, with M.W. additionally contributing to the development and expansion of Section 6. All authors reviewed and approved the final version of the manuscript.
 
\section{Funding}
  M.W. acknowledges funding by the European Union's Horizon Europe programme under grant agreement No. 101215206 (UNCAN-CONNECT).

\bibliography{sn-bibliography}% common bib file

\end{document}